\documentclass[runningheads]{llncs}
\usepackage[T1]{fontenc}
\usepackage{graphicx,verbatim,amsmath}
\usepackage{array}
\usepackage{hyperref}
\usepackage{xurl}
\usepackage{color}

\begin{document}
\title{Healthy Counterfactual Generation via Diffusion Inpainting for Mammography Classification} 
\titlerunning{Healthy Counterfactual Generation for Mammography Classification}
%
%
\author{Inês Cruchinho Garcia \and
Mariana Mourão \and
Francisco Maria Calisto \and
Carlos Santiago \and
Jacinto Nascimento }
\authorrunning{I. Garcia et al.}
%
\institute{
Institute for Systems and Robotics, Instituto Superior Técnico, Lisbon, Portugal\\
\email{ines.cruchinho.garcia@tecnico.ulisboa.pt}
}

\maketitle              
\begin{abstract}
False negatives remain a critical limitation of computer-aided diagnosis (CAD) systems for breast cancer screening due to delayed detection and treatment. To address this issue, we propose a counterfactual data augmentation strategy that generates healthy mammograms by "erasing" lesions from anomalous images, thereby enriching the training distribution.
We train a Denoising Diffusion Probabilistic Model on BI-RADS 1 (healthy) mammograms and use a RePaint-based sampling strategy to inpaint realistic normal tissue within annotated lesion bounding boxes. The resulting healthy counterfactuals replace annotated lesion regions with realistic healthy tissue while preserving patient-specific anatomical structure, as supported by similarity metrics between real and generated images. Image realism was further assessed by radiologists and found to be consistent with the original dataset quality.
We evaluate counterfactual augmentation across four representative classifier architectures: a convolutional neural network (ConvNeXt), a vision transformer (ViT), a vision-language model pre-trained on mammogram-report pairs (Mammo-CLIP) and a multi-scale attention-based multiple-instance learning framework (FPN-MIL). Experiments conducted on the VinDr-Mammo dataset show improvements in sensitivity across all architectures, particularly at 80\% fixed specificity, contributing towards more reliable CAD systems for breast cancer. Code is available at:
\url{https://github.com/ines03garcia/diffusion-based-counterfactual-generation}.


\end{abstract}
\section{Introduction}
False negatives in breast cancer detection systems are associated with delayed diagnosis and worse patient outcomes. Although mammography remains the primary modality for early breast cancer screening, its sensitivity is affected by multiple factors, including inter-patient anatomical variability \cite{Attallah2025ThermogramsCAD}.
Beyond missed detections, an additional concern in computer-aided diagnosis (CAD) is whether predictions are driven by clinically meaningful evidence. Deep learning models may achieve strong aggregate performance while relying on spurious correlations rather than lesion-specific cues \cite{Li2025LetSamplesSpeak}. Such effects have been observed in breast imaging settings \cite{WonJon_SpurBreast_MICCAI2025}. For clinical deployment, CAD systems must not only improve sensitivity but also ensure that predictions are based on regions of interest (ROI) \cite{Saporta2022BenchmarkingSaliency}.

Counterfactuals represent 'what-if' scenarios, such as: how would this patient's mammogram look if there were no findings? Diffusion models have \allowbreak emerged as a leading approach for high-fidelity image generation \cite{Dhariwal2021DiffusionBeatsGANs}. Building on the Denoising Diffusion Probabilistic Model (DDPM) framework, these methods iteratively learn to reverse a gradual noise corruption process, enabling the sampling of images from random Gaussian noise. Commonly used to synthesize medical images \cite{ZhaYou_HighFidelity_MICCAI2025}, including X-rays \cite{XieChu_SVDRR_MICCAI2025} and breast MRIs \cite{Ibarra2025BreastDiffusion}, diffusion models have been recently explored for mammogram generation \cite{Na2024RadiomicsFillMammo}.
Focusing on counterfactuals, their applications include data augmentation \cite{HeoCha_Semantic_MICCAI2025}, explainability \cite{fathi2024decodex} and classifier performance enhancement \cite{DreDav_MeDi_MICCAI2025}, but a critical challenge remains the unintended modification of unrelated morphology \cite{MinHyu_InstructX2X_MICCAI2025}. To address this, our approach restricts modifications to previously annotated lesion regions by employing masked diffusion inpainting. The inpainted area is iteratively conditioned on the surrounding tissue, ensuring anatomical coherence while preventing unintended modifications outside the annotated bounding box. Rather than generating lesions, an approach hindered by data scarcity, high anomaly variability, and complex distributional patterns \cite{ZhaHao_Paired_MICCAI2025}, we generate healthy counterfactuals, i.e., anatomically plausible normal counterparts of originally abnormal images.
Although healthy counterfactuals have previously been explored in mammography \cite{Sanchez2022WhatIsHealthy}, their use as a data augmentation strategy remains largely underexplored. By incorporating these counterfactuals into the training set of four diverse classifier architectures, we aim to enhance the robustness of CAD systems for breast cancer detection. Specifically, we select architectures with distinct inductive biases: a convolutional neural network (ConvNeXt) \cite{Liu2022ConvNeXt}, a transformer-based vision model (ViT) \cite{Dosovitskiy2021ViT}, a mammography-specific image encoder (Mammo-CLIP) \cite{VinDr-Mammo-Clip} and a multiple-instance learning classifier (FPN-MIL) \cite{MouMar_Multiscale_MICCAI2025}.

Comparative evaluation on the VinDr-Mammo dataset against standard augmentation baselines demonstrates consistent sensitivity gains, particularly at a clinically relevant operating point of 80\% specificity, supporting healthy counterfactual augmentation as a strategy for reducing missed detections in mammography classification.

\section{Method}
\begin{figure}[t]
  \centering
  \includegraphics[width=\textwidth]{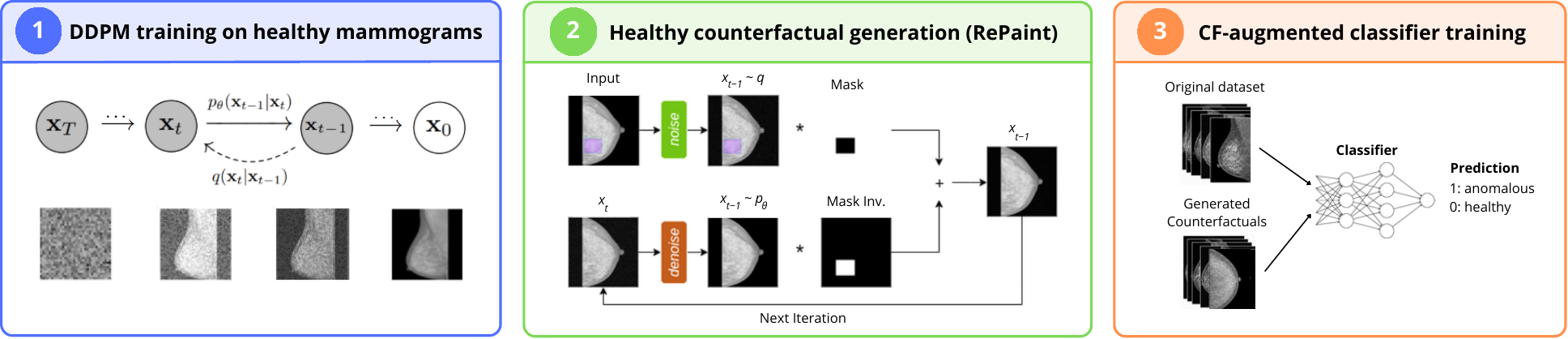}
  \caption{Overview of the proposed pipeline. First, a DDPM is trained on healthy mammograms to learn the distribution of healthy breast tissue. Second, RePaint is used to generate healthy counterfactuals via mask-guided inpainting of lesion regions, using binary masks derived from bounding-box annotations. Finally, the generated counterfactuals are incorporated into the training set for classification.}
  \label{fig:pipeline-overview}
\end{figure}

To investigate whether counterfactuals enhance deep learning classifiers for mammography, we propose a three-stage pipeline (Figure~\ref{fig:pipeline-overview}). First, a DDPM \cite{DDPM} is trained exclusively on BI-RADS~1 mammograms to model the distribution of normal breast tissue. Second, the RePaint~\cite{RePaint} sampling mechanism is employed to inpaint healthy tissue within annotated lesion bounding boxes, producing healthy counterfactuals. Third, the synthetic images are incorporated into the training set to obtain counterfactual-augmented models.

\subsection{Denoising Diffusion Probabilistic Model (DDPM)}
We train a DDPM~\cite{DDPM} with a U-Net backbone to model the distribution of healthy mammograms (BI-RADS~1). Following the standard diffusion framework, Gaussian noise is progressively added to images according to a predefined variance schedule, defining a forward noising process. The network is trained to predict the injected noise at each timestep by minimizing a mean squared error (MSE) objective, thereby learning the corresponding denoising process. In inference, sampling is performed by iteratively denoising random Gaussian noise using the learned reverse transitions. Training exclusively on healthy images enables the model to capture normal breast anatomy for subsequent counterfactual generation.

\subsection{RePaint Inpainting}
We use RePaint~\cite{RePaint} to perform mask-conditioned diffusion inpainting. Given a binary lesion mask $M$, known pixels are re-noised to the current timestep, while masked regions are updated via the learned reverse transition. At each step:
\[
x_{t-1} = M \odot x^{\text{known}}_{t-1} + (1-M) \odot \hat{x}_{t-1}.
\] where $x_{t-1}$ denotes the resulting image at timestep $t-1$, $x^{\mathrm{known}}_{t-1}$ represents the region to preserve with corresponding noise at $t-1$, and $\hat{x}_{t-1}$ denotes the newly generated content based on $x_{t}$. The operator $\odot$ represents element-wise multiplication.

\section{Experiments}

\subsection{Dataset}
We conduct experiments on the publicly available VinDr-Mammo dataset~\cite{VinDr-Mammo}, preprocessed by~\cite{VinDr-Mammo-Clip}. The dataset comprises 5,000 full-field digital mammography exams, each containing four standard views: craniocaudal (CC) and mediolateral oblique (MLO) views of both breasts, totalling 20,000 mammograms. Following the original data splits and folds, 16,000 images (80\%) were used for training and 4,000 (20\%) for testing.
To balance anatomical fidelity and computational efficiency during counterfactual generation, mammograms were resized to $512 \times 512$ pixels using zero padding to preserve aspect ratio. Synthesized images were later resized back to their original resolution of $1520 \times 912$ pixels. Contrast Limited Adaptive Histogram Equalization (CLAHE) was applied to normalize local contrast across mammograms.

\subsection{Counterfactual Generation}

\subsubsection{DDPM Training}
The DDPM was trained on BI-RADS~1 mammograms using an NVIDIA A100 GPU with 80~GB of memory. Although BI-RADS~2 corresponds to benign findings, these cases were excluded from this training stage to avoid generation of any lesion-like structures. We used the OpenAI \texttt{guided-diffusion} implementation in PyTorch. The model is based on a U-Net backbone with 256 base channels and was trained on single-channel images resized to $512 \times 512$ pixels. Training used an MSE noise-prediction objective with a linear variance schedule over 1,000 diffusion timesteps, with an AdamW optimizer ($\mathrm{LR}=1 \times 10^{-4}$, batch size=16), for approximately 8,000 iterations until validation loss convergence.

\subsubsection{Sampling using RePaint}

The trained DDPM was integrated into the RePaint mechanism to generate lesion-free counterfactual mammograms. For each anomalous image, a binary lesion mask was constructed from the annotated bounding box, with lesion pixels set to 0 (to be inpainted) and surrounding tissue set to 1 (to be preserved).

Sampling was performed using mask-conditioned reverse diffusion with 500 respaced timesteps to improve computational efficiency. At each timestep, known regions were re-noised and reinserted to preserve anatomical context, while masked regions were updated according to the learned reverse transition. One counterfactual image was generated per anomalous annotated mammogram and subsequently added to the training set.

\subsection{Counterfactual Augmentation}
To evaluate the impact of counterfactual augmentation across different inductive biases, we selected four classifiers: ConvNeXt \cite{Liu2022ConvNeXt} as a hierarchical CNN, ViT \cite{Dosovitskiy2021ViT} for modelling long-range dependencies, Mammo-CLIP \cite{VinDr-Mammo-Clip} as a VinDr-Mammo pretrained encoder, and FPN-MIL~\cite{MouMar_Multiscale_MICCAI2025} for weakly supervised multi-scale aggregation. 

ConvNeXt and ViT were initialized with ImageNet-pretrained weights, while Mammo-CLIP and FPN-MIL used EfficientNet backbones pretrained on VinDr-Mammo. Models were fine-tuned using binary cross-entropy loss, AdamW ($\mathrm{LR}=3 \times 10^{-4}$), a batch size of 8, cosine annealing and positive class weighting to account for the imbalance between healthy and anomalous samples.

Models were first trained using 4-fold cross-validation to estimate an appropriate training duration. Three folds of the training set were used for model fitting, while the remaining fold, comprising 3,200 images, was used for validation. The optimal number of epochs was averaged across folds and subsequently used to retrain each model on the full training set. For counterfactual augmentation, one healthy counterfactual was generated for each annotated image, adding 1,410 negative samples to the training set, increasing the size of the negative class by 13\%. Positive-class weights were recalculated after augmentation to account for the resulting change in class distribution.

\section{Results}
We evaluated the realism and clinical plausibility of the generated counterfactuals through distributional metrics and radiologists' studies. The effectiveness of the proposed data augmentation is assessed using standard classification performance metrics. We did not perform matched-pair prediction-consistency analysis because its interpretation of what the classifier's learning is limited in the context of counterfactual image generation.

\subsection{Healthy Counterfactual Generation Quality Assessment}
\subsubsection{Distributional Metrics}
We evaluate the realism of the generated images using distributional metrics based on the Fréchet Inception Distance (FID)~\cite{DBLPjournalsHeuselRUNKH17} and two variants of the Fréchet Radiomic Distance (FRD)~\cite{Konz_2026}, which assess perceptual quality and radiomic consistency.

Table~\ref{tab:frd_comparison} shows the results obtained with our approach, where the low FID and FRD values suggest that the edits remain close to the distribution of real healthy mammograms. We also report results from MammoFlow~\cite{du2026mammoflowmultiviewmammogramsynthesis} on VinDr-Mammo, although their results are for full image generation and are not restricted to healthy cases.

We also assess only the generated patch by cropping each counterfactual to the edited bounding-box region and compare the generated healthy patches with real healthy patches from VinDr-Mammo. Our scores are better than those reported by Osuala et al.~\cite{osuala2024enhancingutilityprivacypreservingcancer}, who evaluate synthetic mass patches against real mass patches from the CBIS-DDSM~\cite{SawyerLee2016CBISDDSM}. Although the datasets and patch types differ, our results indicate competitive realism.

\begin{table}[ht]
\centering
\caption{
Quantitative comparison of image realism, measured by FID, and radiomic consistency, measured by FRD, between generated and real mammograms. Lower values indicate greater similarity to the corresponding real-image distribution. Results from the proposed method are shown in bold. Prior results are included for reference, although they are not directly comparable due to differences in protocol.
}
\label{tab:frd_comparison}
\resizebox{\columnwidth}{!}{
\begin{tabular}{lc
>{\centering\arraybackslash}p{1.3cm}
>{\centering\arraybackslash}p{1.3cm}
>{\centering\arraybackslash}p{1.6cm}
>{\centering\arraybackslash}p{1.6cm}}
\hline
\textbf{Compared Distributions} & \textbf{Dataset} & \textbf{Scale} & \textbf{FID$\downarrow$} & \textbf{FRDv0$\downarrow$} & \textbf{FRDv1$\downarrow$} \\ 
\hline
Healthy CF. vs. Real Healthy (ours) & VinDr-Mammo & image & \textbf{10.1} &	\textbf{12.3} & \textbf{7.5} \\
Synthetic vs. Real \cite{du2026mammoflowmultiviewmammogramsynthesis} & VinDr-Mammo & image & 67.5 & -- & 12.4 \\
\hline
Healthy CF. vs. Real Healthy (ours) & VinDr-Mammo & patch & \textbf{37.7} &	\textbf{12.3} & \textbf{10.9} \\
Synthetic Masses vs. Real Masses~\cite{osuala2024enhancingutilityprivacypreservingcancer} & CBIS-DDSM & patch & 58.0 & 18.1 & -- \\
\hline
\end{tabular}
}
\end{table}

\subsubsection{Expert Reader Study}
To complement the metrics-based evaluation, we conducted two expert assessments of the generated healthy counterfactual mammograms: a blind review by an experienced radiologist and a broader inspection of counterfactuals involving multiple radiologists.

\paragraph{Blind Evaluation Test:}
An experienced radiologist with over 35 years of experience blindly evaluated 70 healthy counterfactual mammograms and 30 real healthy mammograms. Each image was rated on a five-point Likert scale in two stages: first for full-image realism, assessing whether the image could plausibly resemble a real mammogram, and then for regional healthiness, assessing whether the tissue within the bounding box appeared healthy.

\begin{figure}[t]
  \centering
  \includegraphics[width=0.6\textwidth]{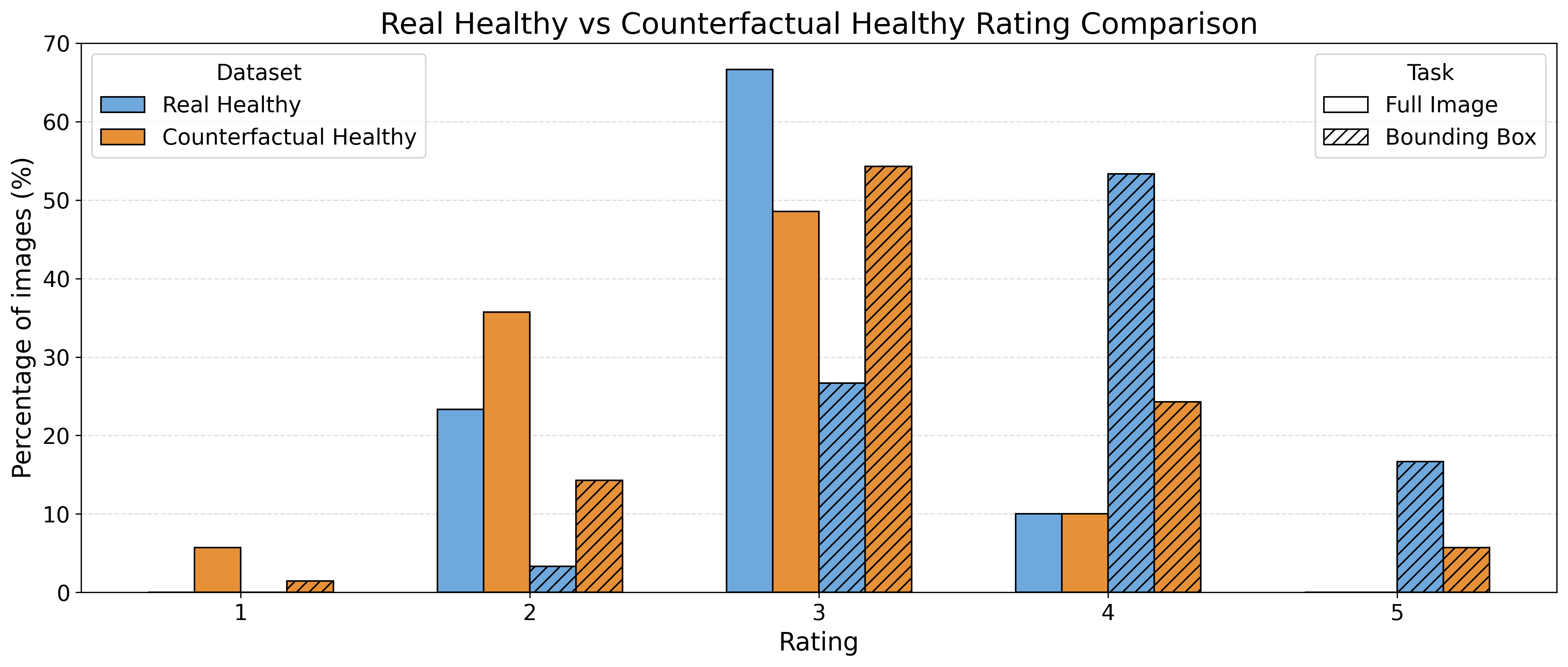}
  \caption{Distribution of radiologist ratings for real healthy mammograms and generated healthy counterfactuals. Ratings were assigned on a five-point Likert scale. For the full-image assessment, higher scores indicated that an image was more likely to resemble a real mammogram; for the bounding-box assessment, higher scores indicated that the highlighted tissue was more likely to resemble healthy breast tissue.}
  \label{fig:rating_percentages}
\end{figure}

The rating distributions are shown in Fig.~\ref{fig:rating_percentages}. For full-image realism, both real and counterfactual mammograms were frequently assigned intermediate ratings (2--3). Counterfactuals showed a slight shift toward lower ratings, indicating a small reduction in perceived realism. A similar trend was observed in the bounding-boxes healthiness assessment: regions from real healthy mammograms received higher ratings than counterfactual regions, although both distributions remained concentrated within a similar range (3--4).

The radiologist also noted several quality limitations in the original dataset, including low contrast, unrealistic soft-tissue appearance, implausible superposition in the axillary region and incomplete breast coverage, with some mammograms cropping posterior or inferior tissue. These observations suggest that part of the perceived lack of realism may reflect limitations already present in the source images rather than artifacts introduced by counterfactual generation.

These findings indicate that the counterfactual edits introduced some perceptible differences, but their ratings remain broadly aligned with those of real healthy mammograms.

\begin{figure}[t]
  \centering
  \includegraphics[width=0.7\textwidth]{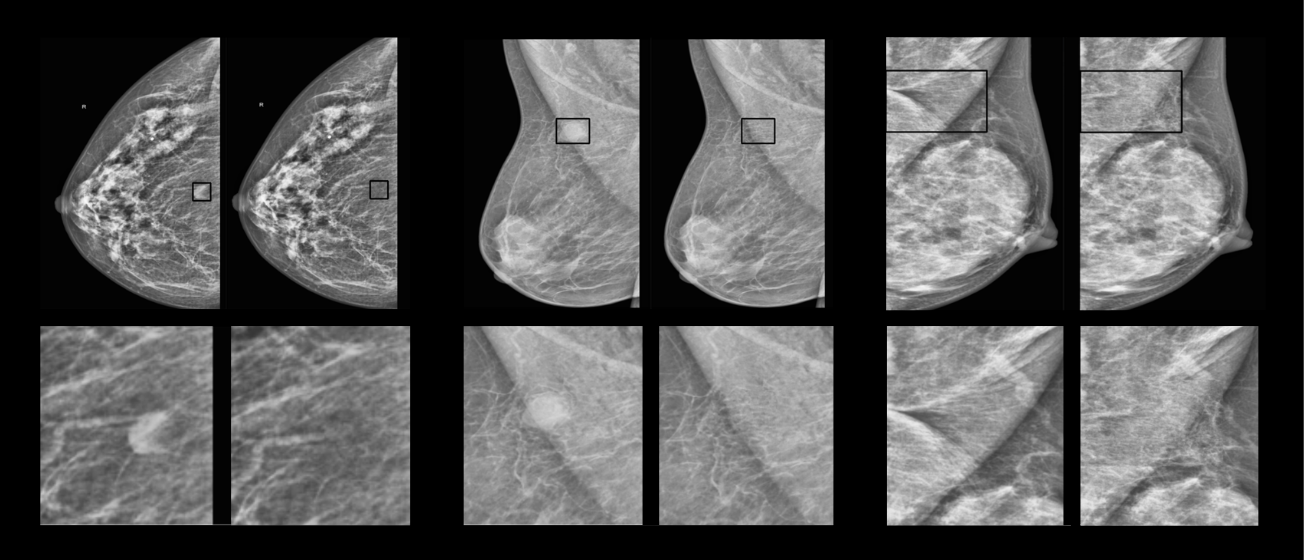}
  \caption{Three lesion-free counterfactual mammograms rated 5/5 for realism by all three reviewing radiologists. For each example, the original image is shown on the left and the corresponding counterfactual on the right. Zoomed views of the regions of interest are shown in the bottom row.}
  \label{fig:inpainting_examples}
\end{figure}

\paragraph{CF-focused assessment:}
A balanced subset of 100 synthetic images, spanning different breast density and BI-RADS categories, was evaluated by six radiologists. Each expert independently reviewed 40 full-image counterfactuals, such that each image received two to three evaluations.
Figure~\ref{fig:inpainting_examples} presents three high-quality counterfactuals that received unanimous positive ratings (5/5) from all reviewing radiologists.

\begin{figure}[t]
    \centering
    \begin{minipage}{0.40\textwidth}
        \centering
        \includegraphics[width=\textwidth]{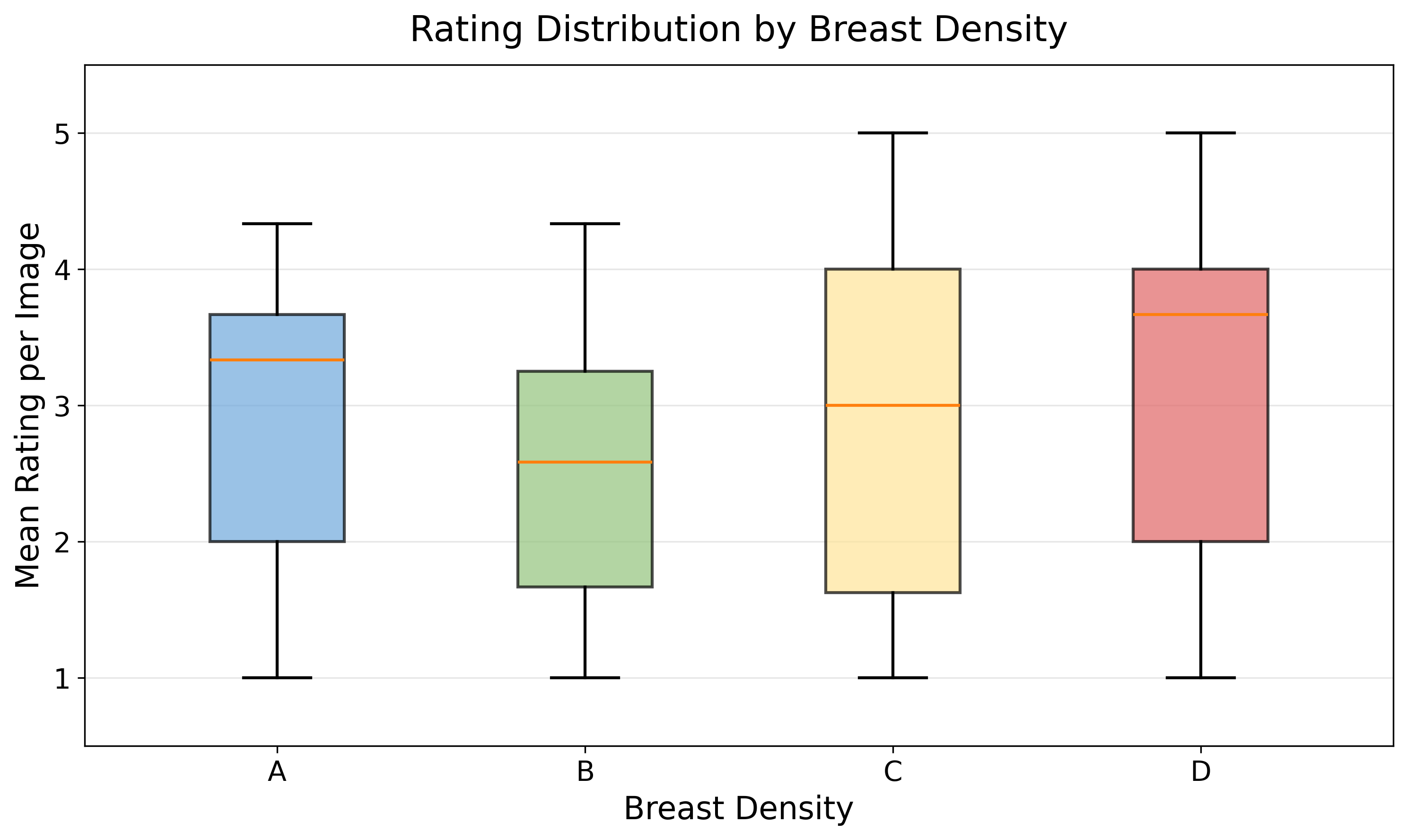}
    \end{minipage}
    \hspace{0.05\textwidth}
    \begin{minipage}{0.40\textwidth}
        \centering
        \includegraphics[width=\textwidth]{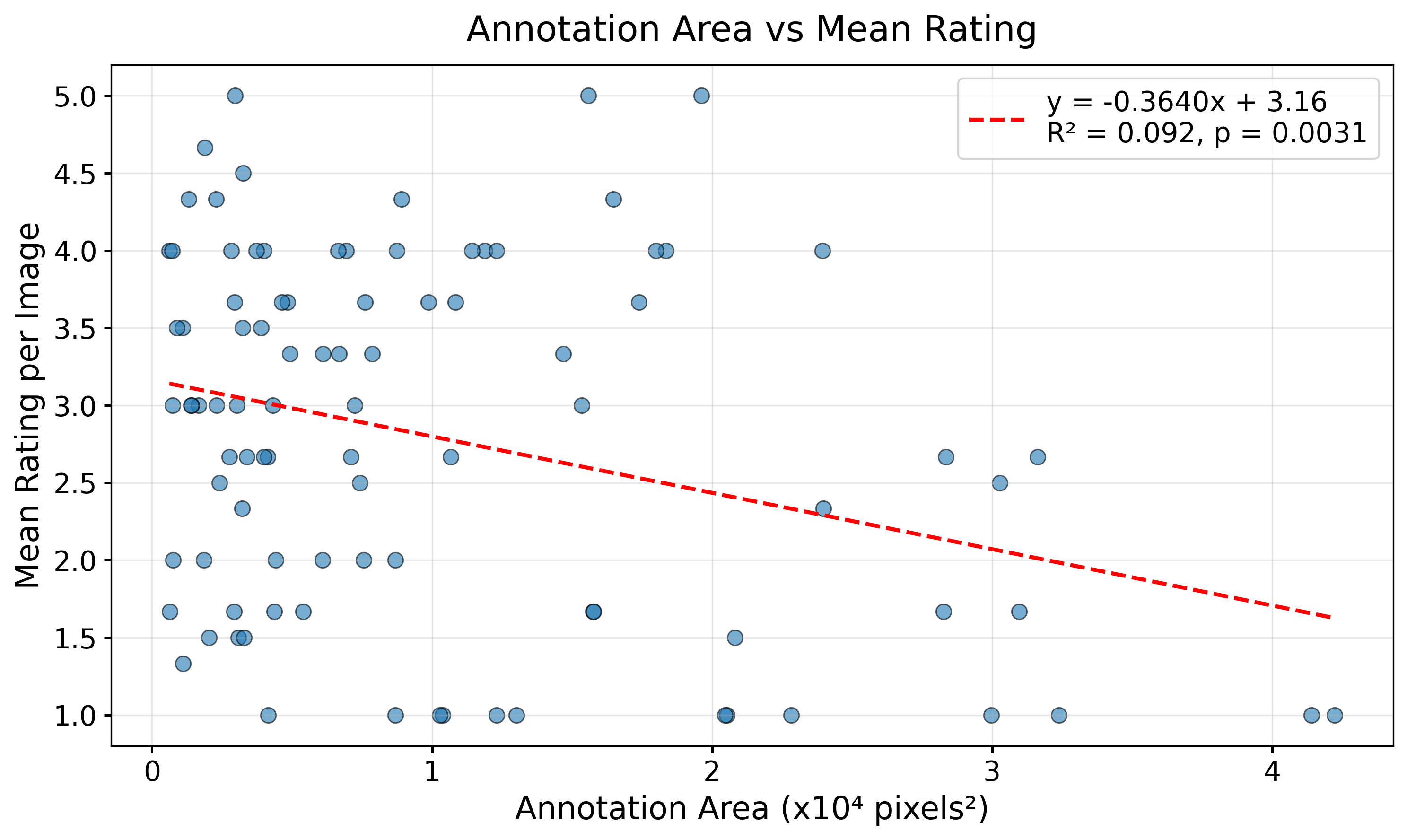}
    \end{minipage}
    \caption{Radiologist evaluation of counterfactual realism for a balanced subset of counterfactuals. Left: Boxplot showing the distribution of mean realism ratings across breast density categories. Right: Scatter plot with linear regression illustrating the correlation between mean realism ratings and annotation area.}
    \label{fig:cf_quality_analysis}
\end{figure}

The rating distributions stratified by breast density are shown in Fig.~\ref{fig:cf_quality_analysis}. Density D mammograms received the strongest overall ratings, while density C cases showed the largest proportion of upper-end ratings compared with densities A and B. This may be partly related to the greater prevalence of denser mammograms in the training dataset.

Inpainting quality was also negatively correlated with lesion size (Fig.~\ref{fig:cf_quality_analysis}). Smaller lesions were reconstructed more convincingly, whereas larger masked regions tended to receive lower ratings.

\subsection{Data Augmentation}
\subsubsection{Classification Performance}
Table~\ref{tab:classification_results} shows the effect of adding healthy counterfactual images as data augmentation across four classification backbones.
Recall and performance at fixed specificity (shown through Recall at 80\% Spec.) improve consistently across all models. The largest recall gains are observed for ConvNeXt and FPN-MIL, with improvements of 4.6\% and 1.9\%, respectively. Improvements in recall at 80\% specificity range from 0.6\% to 1.3\%, suggesting that counterfactual augmentation can improve sensitivity under a clinically relevant operating constraint. Balanced accuracy and F1-score show modest improvements across all models.

Specificity and ROC-AUC reveal a more mixed pattern. While specificity improves slightly for ViT, it decreases for ConvNeXt, Mammo-CLIP, and FPN-MIL. ROC-AUC improves modestly for ViT and Mammo-CLIP, but remains nearly unchanged for ConvNeXt and FPN-MIL.

These results suggest that healthy counterfactual augmentation consistently improves sensitivity, particularly at the clinically relevant operating point of 80\% specificity, while introducing a slight reduction in specificity for some models. Although gains in balanced accuracy, F1-score, and ROC-AUC are more modest, the improved sensitivity highlights the potential of the proposed augmentation strategy for computer-aided breast cancer screening, where reducing missed diagnoses is often prioritised.

\begin{table}[t]
\caption{Classification performance of ConvNeXt, ViT, Mammo-CLIP and FPN-MIL under baseline training and counterfactual augmentation (CF) on the original test set. Improved results are highlighted in bold. Recall values with fixed specificity (80\%) are also reported.}
\centering
\resizebox{\columnwidth}{!}{%
\begin{tabular}{l|cc|cc|cc|cc}
\hline
 & \multicolumn{2}{c|}{ConvNeXt \cite{Liu2022ConvNeXt}} 
 & \multicolumn{2}{c|}{ViT \cite{Dosovitskiy2021ViT}}
 & \multicolumn{2}{c|}{Mammo-CLIP \cite{VinDr-Mammo}}
 & \multicolumn{2}{c}{FPN-MIL \cite{MouMar_Multiscale_MICCAI2025}} \\
 
\textbf{Metric} 
 & Baseline & CF 
 & Baseline & CF 
 & Baseline & CF
 & Baseline & CF \\
\hline

Balanced Acc. & 73.1 & \textbf{73.2} & 72.3 & \textbf{72.7}
              & 77.1 & \textbf{77.3} & 76.7 & \textbf{76.8} \\

Recall        & 63.5 & \textbf{68.1} & 63.1 & \textbf{63.7}
              & 69.0 & \textbf{70.7} & 73.1 & \textbf{75.0} \\

Specificity   & \textbf{82.6} & 78.2 & 81.6 & \textbf{81.7}
              & \textbf{85.2} & 83.8 & \textbf{80.2} & 78.6 \\

F1-score      & 63.9 & \textbf{64.1} & 62.9 & \textbf{63.3}
              & 69.3 & \textbf{69.4} & 68.6 & \textbf{68.7} \\

ROC-AUC       & \textbf{80.3} & 80.2 & 77.8 & \textbf{78.4}
              & 84.6 & \textbf{84.8} & \textbf{84.1} & 84.0 \\

Recall (at 80\% Spec.) & 65.9 & \textbf{66.5}
                       & 63.7 & \textbf{65.0}
                       & 73.7 & \textbf{74.7}
                       & 73.2 & \textbf{74.5} \\
\hline
\end{tabular}%
}
\label{tab:classification_results}
\end{table}

\section{Conclusion}
In this work, we introduced a diffusion-based counterfactual augmentation framework for mammography classification. The proposed method generates lesion-free counterfactuals by training a DDPM exclusively on healthy mammograms and applying RePaint-based inpainting within annotated lesion regions, while preserving the surrounding patient-specific anatomy. 

Distributional metrics and expert reader assessments indicate that the generated images are consistent with real healthy mammograms. When used for data augmentation, they consistently improved sensitivity, particularly at a clinically relevant operating point of 80\% specificity, across diverse classification architectures. Improvements in F1-score, balanced accuracy and ROC-AUC were more modest, while specificity slightly decreased for some models.

Our work proposes healthy counterfactual generation as a strategy for data augmentation in mammography, producing realistic patient-specific lesion-free mammograms that consistently improve sensitivity across models with different inductive biases.

    

\begin{credits}
\subsubsection{\ackname}
We gratefully acknowledge Dr Cristina Ribeiro da Fonseca and
Dr João Abrantes and his team for their assistance in evaluating the
synthetic mammograms in the counterfactual quality assessment.
This work is funded by LARSyS FCT funding
(10.54499/LA/\allowbreak P/\allowbreak 0083/\allowbreak 2020,
10.54499/UIDP/\allowbreak 50009/\allowbreak 2020, and
10.54499/\allowbreak UIDB/\allowbreak 50009/\allowbreak 2020),
and by the FCT project MIA-BREAST
(10.54499/2022.0485.\allowbreak PT \allowbreak DC).

\subsubsection{\discintname}
The authors have no competing interests to declare that
are relevant to the content of this article.
\end{credits}

%
%
%

\bibliographystyle{splncs04}
\bibliography{mybibliography}

@article{Attallah2025ThermogramsCAD,
  author = {Attallah, O.},
  title = {A Deep Learning-Driven CAD for Breast Cancer Detection via Thermograms: A Compact Multi-Architecture Feature Strategy},
  journal = {Applied Sciences},
  year = {2025},
  volume = {15},
  number = {13},
  pages = {7181},
  doi = {10.3390/app15137181}
}

@inproceedings{Li2025LetSamplesSpeak,
  author = {Li, Weiwei and others},
  title = {Let Samples Speak: Mitigating Spurious Correlation by Exploiting the Clusterness of Samples},
  booktitle = {Proceedings of the IEEE/CVF Conference on Computer Vision and Pattern Recognition (CVPR)},
  year = {2025},
  pages = {15486--15496}
}

@inproceedings{WonJon_SpurBreast_MICCAI2025,
  author = {Won, Jong Bum and others},
  title = {{ SpurBreast: A Curated Dataset for Investigating Spurious Correlations in Real-world Breast MRI Classification }},
  booktitle = {proceedings of Medical Image Computing and Computer Assisted Intervention -- MICCAI 2025},
  year = {2025},
  volume = {LNCS 15975},
  pages = {555 -- 564}
}

@article{Saporta2022BenchmarkingSaliency,
  author = {Saporta, Adriel and others},
  title = {Benchmarking Saliency Methods for Chest X-Ray Interpretation},
  journal = {Nature Machine Intelligence},
  year = {2022},
  volume = {4},
  number = {10},
  pages = {867--878},
  doi = {10.1038/s42256-022-00536-x}
}

@inproceedings{HeoCha_Semantic_MICCAI2025,
  author = {Heo, Chanyeong and others},
  title = {{ Semantic Interpolative Diffusion Model: Bridging the Interpolation to Masks and Colonoscopy Image Synthesis for Robust Generalization }},
  booktitle = {proceedings of Medical Image Computing and Computer Assisted Intervention -- MICCAI 2025},
  year = {2025},
  volume = {LNCS 15970},
  pages = {519 -- 529}
}

@inproceedings{fathi2024decodex,
  title = {DeCo{DE}x: Confounder Detector Guidance for Improved Diffusion-based Counterfactual Explanations},
  author = {Nima Fathi and others},
  booktitle = {Medical Imaging with Deep Learning},
  year = {2024},
  url = {https://openreview.net/forum?id=M6CfJ5H7XH}
}

@inproceedings{DreDav_MeDi_MICCAI2025,
  author = {Drexlin, David Jacob and others},
  title = {{ MeDi: Metadata-Guided Diffusion Models for Mitigating Biases in Tumor Classification }},
  booktitle = {proceedings of Medical Image Computing and Computer Assisted Intervention -- MICCAI 2025},
  year = {2025},
  volume = {LNCS 15973},
  pages = {379 -- 388}
}

@inproceedings{MinHyu_InstructX2X_MICCAI2025,
  author = {Min, Hyungi and others},
  title = {{ InstructX2X: An Interpretable Local Editing Model for Counterfactual Medical Image Generation }},
  booktitle = {proceedings of Medical Image Computing and Computer Assisted Intervention -- MICCAI 2025},
  year = {2025},
  volume = {LNCS 15975},
  pages = {279 -- 288}
}

@inproceedings{Sanchez2022WhatIsHealthy,
  author = {Sanchez, Pedro and others},
  title = {What Is Healthy? Generative Counterfactual Diffusion for Lesion Localization},
  booktitle = {Deep Generative Models -- MICCAI 2022 Workshop},
  pages = {34--44},
  year = {2022},
  doi = {10.1007/978-3-031-18576-2_4}
}

@inproceedings{ZhaHao_Paired_MICCAI2025,
  author = {Zhang, Haoxuan and others},
  title = {{ Paired Image Generation with Diffusion-Guided Diffusion Models }},
  booktitle = {proceedings of Medical Image Computing and Computer Assisted Intervention -- MICCAI 2025},
  year = {2025},
  volume = {LNCS 15963},
  pages = {371 -- 381}
}

@inproceedings{Dhariwal2021DiffusionBeatsGANs,
  author = {Dhariwal, Prafulla and others},
  title = {Diffusion Models Beat GANs on Image Synthesis},
  booktitle = {Advances in Neural Information Processing Systems (NeurIPS)},
  year = {2021},
  volume = {34},
  pages = {8780--8794}
}

@inproceedings{XieChu_SVDRR_MICCAI2025,
  author = {Xie, Chun and others},
  title = {{ SV-DRR: High-Fidelity Novel View X-Ray Synthesis Using Diffusion Model }},
  booktitle = {proceedings of Medical Image Computing and Computer Assisted Intervention -- MICCAI 2025},
  year = {2025},
  volume = {LNCS 15963},
  pages = {572 -- 582}
}

@inproceedings{ZhaYou_HighFidelity_MICCAI2025,
  author = {Zhang, Youjian and others},
  title = {{ High-Fidelity Unified One-to-Many Medical Image Synthesis via Text-Conditioned Latent Diffusion }},
  booktitle = {proceedings of Medical Image Computing and Computer Assisted Intervention -- MICCAI 2025},
  year = {2025},
  volume = {LNCS 15975},
  pages = {258 -- 267}
}

@inproceedings{Ibarra2025BreastDiffusion,
  author = {Ibarra, Sebastian and others},
  title = {Comparing Conditional Diffusion Models for Synthesizing Contrast-Enhanced Breast MRI from Pre-contrast Images},
  booktitle = {Artificial Intelligence and Imaging for Diagnostic and Treatment Challenges in Breast Care (Deep-Breath) -- MICCAI 2025},
  pages = {226--236},
  year = {2025},
  doi = {10.1007/978-3-032-05559-0_23}
}

@inproceedings{MouMar_Multiscale_MICCAI2025,
  author = {Mourão, Mariana and others},
  title = {{ Multi-scale Attention-based Multiple Instance Learning for Breast Cancer Diagnosis }},
  booktitle = {proceedings of Medical Image Computing and Computer Assisted Intervention -- MICCAI 2025},
  year = {2025},
  volume = {LNCS 15974},
  pages = {364 -- 374}
}

@inproceedings{Na2024RadiomicsFillMammo,
  author = {Na, Inye and others},
  title = {RadiomicsFill-Mammo: Synthetic Mammogram Mass Manipulation with Radiomics Features},
  booktitle = {Medical Image Computing and Computer Assisted Intervention -- MICCAI 2024},
  pages = {723--733},
  year = {2024}
}

@inproceedings{DDPM,
  author = {Ho, Jonathan and others},
  title = {Denoising diffusion probabilistic models},
  year = {2020},
  booktitle = {Proceedings of the 34th International Conference on Neural Information Processing Systems}
}

@inproceedings{RePaint,
  author = {Lugmayr, Andreas and others},
  booktitle = {2022 IEEE/CVF Conference on Computer Vision and Pattern Recognition (CVPR)},
  title = {RePaint: Inpainting using Denoising Diffusion Probabilistic Models},
  year = {2022},
  volume = {},
  number = {},
  pages = {11451-11461},
  doi = {10.1109/CVPR52688.2022.01117}
}

@article{VinDr-Mammo,
  author = {Nguyen, Hieu T. and others},
  title = {VinDr-Mammo: A large-scale benchmark dataset for computer-aided diagnosis in full-field digital mammography},
  year = {2022},
  doi = {10.1101/2022.03.07.22272009},
  journal = {medRxiv}
}

@inproceedings{VinDr-Mammo-Clip,
  author = {Ghosh, Shantanu and others},
  title = {{ Mammo-CLIP: A Vision Language Foundation Model to Enhance Data Efficiency and Robustness in Mammography }},
  booktitle = {proceedings of Medical Image Computing and Computer Assisted Intervention -- MICCAI 2024},
  year = {2024},
  volume = {LNCS 15012},
  pages = {632 -- 642}
}

@inproceedings{Dosovitskiy2021ViT,
  author = {Dosovitskiy, Alexey and others},
  title = {An Image is Worth 16x16 Words: Transformers for Image Recognition at Scale},
  booktitle = {International Conference on Learning Representations (ICLR)},
  year = {2021},
  url = {https://openreview.net/forum?id=YicbFdNTTy}
}

@inproceedings{Liu2022ConvNeXt,
  author = {Liu, Zhuang and others},
  title = {A ConvNet for the 2020s},
  booktitle = {Proceedings of the IEEE/CVF Conference on Computer Vision and Pattern Recognition (CVPR)},
  pages = {11966--11976},
  year = {2022},
  doi = {10.1109/CVPR52688.2022.01167}
}

@misc{osuala2024enhancingutilityprivacypreservingcancer,
  title = {Enhancing the Utility of Privacy-Preserving Cancer Classification using Synthetic Data},
  author = {Richard Osuala and others},
  year = {2024},
  eprint = {2407.12669},
  archivePrefix = {arXiv},
  primaryClass = {cs.CV}
}

@misc{du2026mammoflowmultiviewmammogramsynthesis,
  title = {MammoFlow: Multiview Mammogram Synthesis with Anatomically Consistent Flow Matching},
  author = {Yuexi Du and others},
  year = {2026},
  eprint = {2606.28537},
  archivePrefix = {arXiv},
  primaryClass = {cs.CV}
}

@article{DBLPjournalsHeuselRUNKH17,
  author       = {Heusel, Martin and others},
  title        = {GANs Trained by a Two Time-Scale Update Rule Converge to a Nash Equilibrium},
  journal      = {CoRR},
  volume       = {abs/1706.08500},
  year         = {2017},
  url          = {http://arxiv.org/abs/1706.08500},
  eprinttype   = {arXiv},
  eprint       = {1706.08500},
  bibsource    = {dblp computer science bibliography, https://dblp.org}
}

@article{Konz_2026,
   title={Fréchet radiomic distance (FRD): A versatile metric for comparing medical imaging datasets},
   volume={110},
   ISSN={1361-8415},
   url={http://dx.doi.org/10.1016/j.media.2026.103943},
   DOI={10.1016/j.media.2026.103943},
   journal={Medical Image Analysis},
   publisher={Elsevier BV},
   author={Konz, Nicholas and others},
   year={2026},
   month=May, pages={103943}
}

@misc{SawyerLee2016CBISDDSM,
  author    = {Sawyer-Lee, R. and others},
  title     = {Curated Breast Imaging Subset of DDSM (CBIS-DDSM)},
  year      = {2016},
  publisher = {The Cancer Imaging Archive},
  doi       = {10.7937/K9/TCIA.2016.7O02S9CY}
}

\end{document}